\def\year{2020}\relax
\documentclass[letterpaper]{article} 
\usepackage{aaai20}  
\usepackage{times}  
\usepackage{helvet} 
\usepackage{courier}  
\usepackage[hyphens]{url}  
\usepackage{graphicx} 
\usepackage{graphicx}  
\usepackage{array, multirow}
\usepackage{amsfonts}
\usepackage{amsmath}
\usepackage{mathtools}
\usepackage{rotating} 
\usepackage{rotfloat}
\usepackage{url}

\title{Multimodal Transformer with Pointer Network for the DSTC8 AVSD Challenge}
\author{Hung Le$^{1,2}$, Nancy F. Chen$^2$ \\
$^1$Singapore Management University\\
$^2$Institute of Inforcomm Research (I2R), Singapore\\
\url{hungle.2018@phdcs.smu.edu.sg}\\
\url{nfychen@i2r.a-star.edu.sg}
}

\begin{document}

\maketitle

\begin{abstract}
Audio-Visual Scene-Aware Dialog (AVSD) is an extension from Video Question Answering (QA) whereby the dialogue agent is required to generate natural language responses to address user queries and carry on conversations. 
This is a challenging task as it consists of video features of multiple modalities, including text, visual, and audio features. 
The agent also needs to learn semantic dependencies among user utterances and system responses to make coherent conversations with humans. 
In this work, we describe our submission to the AVSD track of the 8th Dialogue System Technology Challenge. 
We adopt dot-product attention to combine text and non-text features of input video. 
We further enhance the generation capability of the dialogue agent by adopting pointer networks to point to tokens from multiple source sequences in each generation step. 
Our systems achieve high performance in automatic metrics and obtain 5th and 6th place in human evaluation among all submissions.
\end{abstract}

\section{Introduction}
\begin{figure}[h]
	\centering
	\resizebox{1.0\columnwidth}{!} {
	\includegraphics{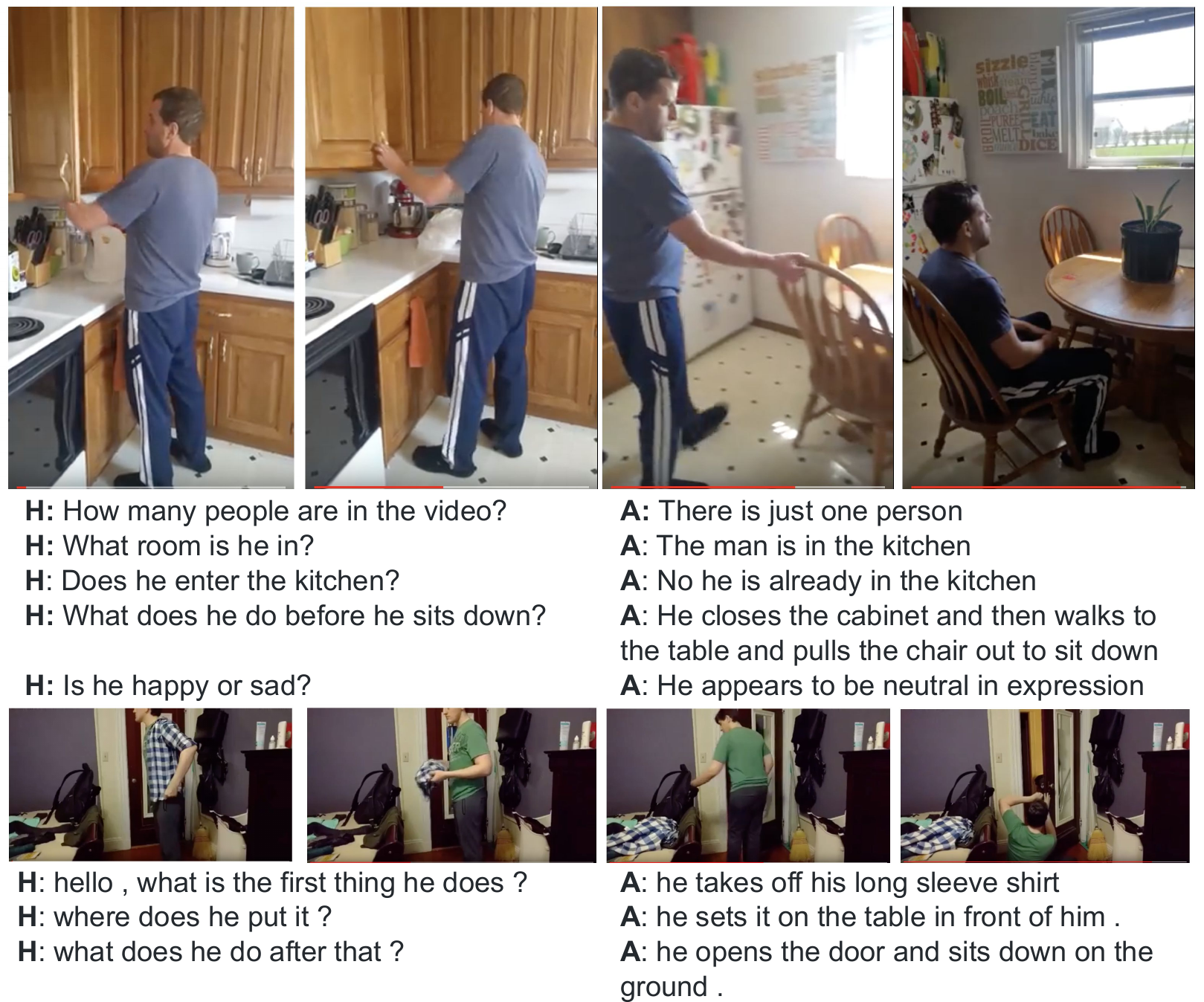}
	}
	\caption{Examples from the AVSD benchmark \cite{alamri2018audio}. Each line includes an utterance from human (\textbf{H}) and the corresponding response from the agent (\textbf{A}).}
	\label{fig:samples}
\end{figure}
AVSD is a challenging task as it involves complex dependencies from features of multiple modalities.
First, the video input typically involves both visual and audio features, each of which contains various information related to the current dialogue context and user utterance.
For example, in Figure \ref{fig:samples}, certain questions from user concern either visual or audio information or both. The two types of features can complement each other to support the dialogue agent to generate responses. 
Second, dialogues also involve complex semantic dependencies among dialogue turns, each consisting of a pair of user utterance and system response. 
For example, in the second turn of the second (lower) example in Figure \ref{fig:samples}, the dialogue agent needs to refer to the previous user utterance and system response to understand what the user is asking. 
We are motivated to address these two challenges by adopting the Multimodal Transformer Network  (MTN) \cite{le-etal-2019-multimodal}.
The model adopts attention mechanism that focuses on the interaction between each token position in text sequences and each temporal step of video visual and audio features. The multi-head structure allows the models to project feature representation to different feature spaces and detect different types of dependencies.  
In addition, while previous work has achieved promising results through complex reasoning neural networks to select important video features \cite{sanabria2019cmu} \cite{hori2019end}, we further investigate the generation capability of the models by using pointer network that can point to tokens from multiple source sequences during each generation step. 
Pointer network is widely used in summarization problem whereby pointer network is used to copy tokens from the source text input to generate summarizing sentences. 
We are motivated by this strategy and adopt into video dialogue task to enhance the quality of the generated system responses. 
We experiment with various model variants and notice interesting findings to improve our model performance.
we noted that adopting pointer network generation can boost performance significantly. 
This could be explained due to the enhanced generation capability of the models being able to copy tokens from relevant text input. 
We present comprehensive experiments and reported results on the validation and test sets which lead to these findings.

\section{Related Work}
The AVSD benchmark in DSTC7 and DSTC8 is considered an extension of two major research directions: Video QA and Dialogues.
Video QA models \cite{xu2017video} \cite{fan2019heterogeneous} \cite{gao2018motion} aim to improve the text and vision reasoning to be able to answer questions from users about a given video.
Compared to Video QA, video dialogues such as the AVSD benchmark, however, has two major challenges:
(1) First, in video dialogues, the model is required to have a strong language understanding over text input, including not only user queries but also past dialogue turns. A user query may refer to previous information mentioned in previous dialogue turns and such references must be learned to answer the query correctly.
(2) Secondly, most of video QA models \cite{jang2017tgif} \cite{lei-etal-2018-tvqa} \cite{kim2019progressive} are more suitable for the retrieval-based setting. In this setting, the model is typically given a list of response candidates and the model has to select one of them as the output. Compared to AVSD, this retrieval-based setting might not be appropriate as dialogue agents need to be able to converse with human users by generating natural responses rather than selecting from a predefined list of sentences.
We are motivated by these two major challenges and propose to improve the language modeling part through pointer networks \cite{nips2015pointer}. 
Adapting to a video dialogue task, we enhance our generative network with pointer distributions over source sequences and construct multiple vocabulary distributions during each generation steps. 

\section{Models}
The input includes a video $V$, dialogue history of $(t-1)$ turns. 
Each turns consists of a pair of (human utterance, sytem response) $(H_1, A_1,..., H_{t-1}, A_{t-1})$, and current human utterance $H_t$. 
The output is a system response $A_t$.
The input video can contain features in different modalities, including vision, audio, and text (such as video caption or subtitle). 
Given current dialogue turn $t$, we can denote each text input as a sequence of tokens.
Each token is represented by a unique token index from a vocabulary set $V$.
We denote the dialogue history $X_{\mathrm{his}}$, user utterance $X_{\mathrm{que}}$, text input of video $X_{\mathrm{cap}}$ 
, and output response $Y$. 

\subsection{Video Features}
Following similar video-based NLP tasks such as video captioning \cite{aafaq2019spatio} and video QA \cite{jang2017tgif}, we assume access to a pretrained model to extract visual or audio features of input video. 
We extracted the visual features from a pretraind 3D-CNN based ResNext \cite{hara2018can} similarly as \cite{sanabria2019cmu}.
The 3D-CNN model extracts the video features based on clips rather than frames.
The clip-based information is expected to be more consistent and less noisy than frame-based information. 
To sample clips, we use a window size of 16 video frames and stride of 16 frames. 
We denote the extracted features as the representation $Z_\mathrm{vis} \in \mathbb{R}^{F \times 2048}$ for visual features, where $F$ is the number of resulting video clips and 2048 is the output dimension in ResNext.
We used the ResNext101 pretrained on the Kinetics Human Action Video benchmark. 
For audio features, following \cite{hori2019end}, we use a pretrained VGGish model \cite{hershey2017cnn}.
This model is based on the image CNN model VGG to extract the temporal variation of video sound.
The output is a 128-dimensional representation. 
We denote the extracted features as the representation $Z_\mathrm{aud} \in \mathbb{R}^{F \times 128}$.

\subsection{Baseline}
The baseline for AVSD benchmark \cite{alamri2018audio} \cite{hori2019end} was provided by the organizers and based on feature fusioning approach proposed by \cite{yu2016video}. 
Video features of multiple modalities, including visual and audio, are combined by passing them through a linear transformation to a common target dimension. The projected representation is used as input to a softmax layer to combine scores of each temporal steps of visual or audio features. 

\subsection{Multimodal Transformer Network (MTN)}
We adopt the MTN model proposed by \cite{le-etal-2019-multimodal} in the AVSD benchmark in DSTC8. 
To improve the performance, we enhance the generation capability of the model and investigate an ensemble approach. 
We summarize the MTN model and our changes below.

\noindent \textbf{Multi-head Attention}. The MTN model adopts the multi-head dot-product attention mechanism proposed by \cite{vaswani17attention} to obtain dependencies between each token in text sequences and temporal variation of video features. 
Specifically, the model considered attention from query to other video feature modalities, including visual and sound features. The output from this attention network is used as input in the decoder. 
The decoder adopts a similar attention mechanism but the attention direction is from the target system responses to other information. 
We denote the attention operation of 2 sequence representations from $Z_1 \in \mathbb{R}^{L_{Z_1} \times d}$ to $Z_2 \in \mathbb{R}^{L_{Z_2} \times d}$ as defined by \cite{vaswani17attention} as:
\begin{align}
    Z_\mathrm{att} &= \mathrm{MultiHeadAtt}(Z_1, Z_2) \in \mathbb{R}^{L_{Z_1}  \times d} \label{eq1} \\
    Z_\mathrm{out} &= \mathrm{LayerNorm}(\mathrm{FF}(Z_\mathrm{att}) + Z_1) \in \mathbb{R}^{L_{Z_1}  \times d} \label{eq2}
\end{align}
where $d$ is the embedding dimension. The attention operation is combined with feed-forward network and skip connection to combine information of the original $Z_1$ with $Z_\mathrm{att}$.
The attention is performed over multiple rounds and in each round, the output is used as input to the next attention steps.
This technique allows progressive feature learning to detect complex dependencies between different information. MTN adopts the Equation \ref{eq1} and \ref{eq2} in query-guided and target-response-guided attention layers to obtain dependencies between user queries/target responses and other input.
First, user query/utterance is used to select important video and audio features of the video. 
For each type of features, the embeddings of user query $Z_\mathrm{que} \in \mathbb{R}^{L_\mathrm{que} \times d}$ is passed to a self-attention layer and another attention layer that attends on video information. Firstly, the query features are used to attend on temporal visual information: 
\begin{align*}
    Z^\mathrm{que2que(1)}_\mathrm{att} &= \mathrm{MultiHeadAtt}(Z_\mathrm{que}, Z_\mathrm{que})  \\
    Z^\mathrm{que2que(1)}_\mathrm{out} &= \mathrm{LayerNorm}(\mathrm{FF}(Z^\mathrm{que2que(1)}_\mathrm{att}) + Z_\mathrm{que}) \\
    Z^\mathrm{que2vis}_\mathrm{att} &= \mathrm{MultiHeadAtt}(Z^\mathrm{que2que(1)}_\mathrm{out}, Z_\mathrm{vis})  \\
    Z^\mathrm{que2vis}_\mathrm{out} &= \mathrm{LayerNorm}(\mathrm{FF}(Z^\mathrm{que2vis}_\mathrm{att}) + Z^\mathrm{que2que(1)}_\mathrm{out}) 
\end{align*}
Each output has the same dimension as $\mathbb{R}^{L_\mathrm{que} \times d}$. Similarly, the query features are used to attend on temporal audio information: 
\begin{align*}
    Z^\mathrm{que2que(2)}_\mathrm{att} &= \mathrm{MultiHeadAtt}(Z_\mathrm{que}, Z_\mathrm{que})  \\
    Z^\mathrm{que2que(2)}_\mathrm{out} &= \mathrm{LayerNorm}(\mathrm{FF}(Z^\mathrm{que2que(2)}_\mathrm{att}) + Z_\mathrm{que}) \\
    Z^\mathrm{que2aud}_\mathrm{att} &= \mathrm{MultiHeadAtt}(Z^\mathrm{que2que(2)}_\mathrm{out}, Z_\mathrm{vis})  \\
    Z^\mathrm{que2aud}_\mathrm{out} &= \mathrm{LayerNorm}(\mathrm{FF}(Z^\mathrm{que2aud}_\mathrm{att}) + Z^\mathrm{que2que(2)}_\mathrm{out}) 
\end{align*}
The self-attention is applied separately for each feature type to allow the model to independently select different information from user query for different types of video features.
The two representations $Z^\mathrm{que2vis}_\mathrm{out}$ and $Z^\mathrm{que2aud}_\mathrm{out}$ contain temporally attended audio and visual features from video. They are passed to the decoder network which processes information from text input (user queries, dialogue history) as well as video input.
Specifically, the target responses $Y$ is embedded into representation $Z_\mathrm{res} \in \mathbb{R}^{L_{Y} \times d}$ and passed to 4 text-to-text attention: self-attention, response-to-dialogue-history attention, and response-to-query-attention, and response-to-caption attention.
\begin{align*}
    Z^\mathrm{res2res}_\mathrm{att} &= \mathrm{MultiHeadAtt}(Z_\mathrm{res}, Z_\mathrm{res})  \\
    Z^\mathrm{res2res}_\mathrm{out} &= \mathrm{LayerNorm}(\mathrm{FF}(Z^\mathrm{res2res}_\mathrm{att}) + Z_\mathrm{res}) \\
    Z^\mathrm{res2his}_\mathrm{att} &= \mathrm{MultiHeadAtt}(Z^\mathrm{res2res}_\mathrm{out}, Z_\mathrm{his}) \\
    Z^\mathrm{res2his}_\mathrm{out} &= \mathrm{LayerNorm}(\mathrm{FF}(Z^\mathrm{res2his}_\mathrm{att}) + Z^\mathrm{res2res}_\mathrm{out}) \\
    Z^\mathrm{res2cap}_\mathrm{att} &= \mathrm{MultiHeadAtt}(Z^\mathrm{res2his}_\mathrm{out}, Z_\mathrm{cap})  \\
    Z^\mathrm{res2cap}_\mathrm{out} &= \mathrm{LayerNorm}(\mathrm{FF}(Z^\mathrm{res2cap}_\mathrm{att}) + Z^\mathrm{res2his}_\mathrm{out}) \\
    Z^\mathrm{res2que}_\mathrm{att} &= \mathrm{MultiHeadAtt}(Z^\mathrm{res2cap}_\mathrm{out}, Z_\mathrm{que})  \\
    Z^\mathrm{res2que}_\mathrm{out} &= \mathrm{LayerNorm}(\mathrm{FF}(Z^\mathrm{res2que}_\mathrm{att}) + Z^\mathrm{res2cap}_\mathrm{out}) 
\end{align*}
The last output is used to attend on the video attended features obtained from query-guided attention layers.
\begin{align*}
    Z^\mathrm{res2aud}_\mathrm{att} &= \mathrm{MultiHeadAtt}(Z^\mathrm{res2que}_\mathrm{out}, Z^\mathrm{que2aud}_\mathrm{out})  \\
    Z^\mathrm{res2aud}_\mathrm{out} &= \mathrm{LayerNorm}(\mathrm{FF}(Z^\mathrm{res2aud}_\mathrm{att}) + Z^\mathrm{res2que}_\mathrm{out}) \\
    Z^\mathrm{res2vis}_\mathrm{att} &= \mathrm{MultiHeadAtt}(Z^\mathrm{res2aud}_\mathrm{out}, Z^\mathrm{que2vis}_\mathrm{out})  \\
    Z^\mathrm{res2vis}_\mathrm{out} &= \mathrm{LayerNorm}(\mathrm{FF}(Z^\mathrm{res2vis}_\mathrm{att}) + Z^\mathrm{res2aud}_\mathrm{out}) 
\end{align*}
The MTN architecture allows the information from different text input and video information from different modalities is incorporated sequentially into the target response representation. 
Adopting the skip connection technique, MTN network can be used to progressively learn and refine signals obtained in each attention steps.
For query-guided attention layers, the progressive learning is done by replacing $Z^\mathrm{que2vis}_\mathrm{out}$ or $Z^\mathrm{que2aud}_\mathrm{out}$ as $Z_\mathrm{que}$ as $Z_\mathrm{que}$ in the next round of attention.
Similarly, in decoder layers, signals can be further attended progressively by replacing $Z^\mathrm{res2vis}_\mathrm{out}$ as $Z_\mathrm{res}$ in the next round of attention.

\noindent \textbf{Pointer Generator}.
We examine an extension of MTN by adopting the pointer network \cite{nips2015pointer} to generate system responses. 
We propose to use pointer network to point to tokens from different input text sequences and construct different vocabulary distribution $P_\mathrm{vocab} \in R^{L_Y \times \|V\|}$ where $V$ is a predefined vocabulary set based on words in the training set.
Given an input text $X$ with embedding representation $Z_{X} \in \mathbb{R}^{L_X \times d}$ and the output from the last attention layer from the decoder $Z^\mathrm{dec}_\mathrm{out} \in \mathbb{R}^{L_Y \times d}$, we construct the pointer distribution by the dot-product attention:
\begin{align}
    P^X_\mathrm{ptr} =  \mathrm{Softmax}(Z^\mathrm{dec}_\mathrm{out} Z^\mathsf{T}_{X}) \in \mathbb{R}^{L_Y \times L_X}
\end{align}
For each position in $L_Y$, the pointer distribution is used to construct a distribution over vocabulary set $V$ where the probability of each token is accumulated from the pointer distribution of the corresponding position. Given a position $i$ in the target response, the vocabulary distribution of this position is defined based on the pointer distribution is defined as:
\begin{align}
    P^{X}_\mathrm{vocab}[i] = \{p(w_j): p(w_j)=\sum P^X_\mathrm{ptr}[i](w_j)\} 
\end{align}
\noindent where $P[i]$ denotes the row $i$ from probability matrix $P$. the We concatenate the probability in all position $i$ to obtain $P^{X}_\mathrm{vocab} \in \mathbb{R}^{L_Y \times \|V\|}$.
For each text input sequence, we obtain the pointer distribution and corresponding vocabulary distribution: $P^\mathrm{his}_\mathrm{vocab}$ for $Z_\mathrm{his}$, $P^\mathrm{cap}_\mathrm{vocab}$ for $Z_\mathrm{cap}$, and $P^\mathrm{que}_\mathrm{vocab}$ for $Z_\mathrm{que}$. 
Besides these pointer-based vocabulary distributions, we adopt a linear transformation layer with Softmax to allow the models to generate tokens not included in any text input sequences.
\begin{align}
    P^\mathrm{gen}_\mathrm{vocab} = \mathrm{Softmax}(Z^\mathrm{dec}_\mathrm{out} W_\mathrm{gen}) \in \mathbb{R}^{L_Y \times \|V\|} \label{eq5}
\end{align}
\noindent where $W_\mathrm{gen} \in {d \times \|V\|}$. To combine the vocabulary distributions, we compute importance scores based on a contextual vectors concatenated from the component input text representations and the output of the decoder.
\begin{align}
    Z_\mathrm{ctx} &= Z^\mathrm{exp}_\mathrm{his} \oplus Z^\mathrm{exp}_\mathrm{que} \oplus Z^\mathrm{exp}_\mathrm{cap} \oplus Z_\mathrm{res} \oplus Z^\mathrm{dec}_\mathrm{out}  \in \mathbb{R}^{L_Y \times 5d}\\
    S &= \mathrm{Softmax}(Z_\mathrm{ctx} W_\mathrm{ctx}) \in \mathbb{R}^{L_Y \times 4}
\end{align}
\noindent where $Z^\mathrm{exp}$ is the expanded version of $Z$ to match the dimensions of $Z_{res} \in \mathbb{R}^{L_Y \times d}$, and $W_\mathrm{ctx} \in \mathbb{R}^{5d \times 4}$. The final vocabulary distribution is computed as the weighted sum of pointer-based distributions and generation-based distribution based on the score matrix $S$. The resulting distribution is denoted as $P_\mathrm{vocab} \in \mathbb{R}^{L_Y \times \|V\|}$.

\noindent \textbf{Optimization}. 
We optimize the model by training it to minimize the generation loss:
\begin{align}
\mathcal{L}_\mathrm{gen} = \sum_{i=0}^{L_Y} -\log(P_\mathrm{vocab}(y_i))
\end{align}
In addition, a key component of the MTN model is the auxiliary loss function applied to the output of query-guided attention. 
This technique was proposed by \cite{le-etal-2019-multimodal} to make the training more stable by using the output of attended features as representations for re-generating the user query. 
This auto-encoder technique was motivated from the multi-task learning approach in neural machine translation (NMT) \cite{luong2016iclr_multi}.
The difference is that MTN extracts the intermediate representations from the (auto-)encoder as video signals for decoding responses rather than just the hidden states of an LSTM encoder of the source sequence in the NMT setting. 
To re-generate user queries, the output from query-guided attention is passed to a linear transformation. We share the weights of the linear layer with $W_\mathrm{gen}$ in Equation \ref{eq5}.
\begin{align}
    P^\mathrm{qae,vis}_\mathrm{vocab} &= \mathrm{Softmax}(Z^\mathrm{que2vis}_\mathrm{out} W_\mathrm{gen}) \in \mathbb{R}^{L_\mathrm{que} \times \|V\|} \\
    P^\mathrm{qae,aud}_\mathrm{vocab} &= \mathrm{Softmax}(Z^\mathrm{que2aud}_\mathrm{out} W_\mathrm{gen}) \in \mathbb{R}^{L_\mathrm{que} \times \|V\|} 
\end{align}
The auto-encoding loss is defined as:
\begin{align}
    \mathcal{L}^\mathrm{vis}_\mathrm{qae} &= \sum_{i=0}^{L_\mathrm{que}} -\log(P^\mathrm{qae,vis}_\mathrm{vocab}(w_i)) \\
    \mathcal{L}^\mathrm{aud}_\mathrm{qae} &= \sum_{i=0}^{L_\mathrm{que}} -\log(P^\mathrm{qae,aud}_\mathrm{vocab}(w_i))
\end{align}
The model are jointly trained with all losses.
\begin{align}
    \mathcal{L} = \mathcal{L}_\mathrm{gen} + \alpha \mathcal{L}^\mathrm{vis}_\mathrm{qae} + \beta \mathcal{L}^\mathrm{aud}_\mathrm{qae}
\end{align}
We simply set $\alpha$ and $\beta$ to 1 for joint training. 

\noindent \textbf{Ensemble Models}. A popular technique to improve the performance is to ensemble models trained in different settings. 
In our submission, we ensemble models trained independently with different video feature types and different feature pretrained models. 
In each model $m$, we obtain the output vocabulary distribution $P^{(m)}_\mathrm{vocab}$. 
The ensembled vocabulary distribution is simply the sum of all vocabulary distributions of component model variants. The resulting summation is passed through a normalization layer to normalize all values from 0 to 1. 
\begin{align}
    P^\mathrm{ensemble}_\mathrm{vocab} = \mathrm{Normalize}(\sum P^{(m)}_\mathrm{vocab}) \in \mathbb{R}^{L_Y \times \|V\|}
\end{align}

\section{Experiments}
\subsection{Dataset}
\noindent We use the AVSD dataset provided in DSTC8 \cite{alamri2018audio} \cite{hori2019end} which contains dialogues grounded on the Charades videos \cite{sigurdsson2016hollywood}.
Following the same track in the DSTC7 challenge, the DSTC8 organizers provided crowd-sourced data of video-based dialogues, including user questions and system responses constructed as dialogues, video captions, and video summaries.
We present a summary of the dataset for training, validation, and test set in Table \ref{tab:datasets}.
The statistics of the official test dataset for DSTC8 challenge are comparable to those in the DSTC7 challenge: 1,710 dialogues and more than 6,700 dialogue turns. 
Please refer to more details on data collection described in \cite{alamri2018audio}.
We construct the vocabulary set $V$ including unique tokens in the training set. 
In our experiments, we use the provided video summary annotation as the video-dependent text input. 
\begin{table}[htbp]
	\centering
	\begin{tabular}{llll}
    \hline
    \textbf{\#}     & \multicolumn{1}{c}{\textbf{Train}} & \multicolumn{1}{c}{\textbf{Val.}} & \multicolumn{1}{c}{\textbf{DTSC7 Test}}\\ \hline
    Dialogs & 7,659                              & 1,787                             & 1,710                    \\ 
                                     Turns   & 153,180                            & 35,740                            & 13,490                     \\ 
                                     Words   & 1,450,754                          & 339,006                           & 110,252                   \\ \hline
    \end{tabular}
	\caption{Summary of DSTC8 AVSD benchmark.
	}
	\label{tab:datasets}
\end{table}

\subsection{Training Procedure}
\noindent We adopt the Adam optimizer \cite{kingma2014adam} with $\beta_1=0.9$, $\beta_2=0.98$, and $\epsilon=10^-9$.
We adopt a learning rate strategy similar to \cite{vaswani17attention}. 
We set the learning rate \textit{warm-up} step to 13,000 training steps and train models up to $50$ epochs. 
We initialize all models with uniform distribution \cite{glorot2010understanding}. 
We select the best models based on the average loss per epoch in the validation set. 
We experiments with following model hyper-parameters: 
embedding dimension $d=512$, 
number of rounds of attention $N=6$, 
attention heads $h_\mathrm{att}=16$.
We tuned hyper-parameters following grid-search over the validation set. 
We allow the pointer generator to point to tokens of video summary and the last user query. 
Experiment results with other combinations of input text sequences for pointer generator are reported in the Ablation Analysis. 
In all experiments with more than one feature type, we adopt the ensemble strategy as described above. 
We select a batch size of 32 and dropout rate of 0.5.
The dropout is applied to all layers except the generator network layers. 
We train our models by applying label smoothing \cite{szegedy2016rethinking} on the target system responses $Y$. 
During inference, we adopt a beam search technique with a beam size of $5$ and a length penalty of $1.0$.

\subsection{Results}
\noindent We report the objective scores, including BLEU \cite{papineni2002bleu}, METEOR \cite{banerjee2005meteor}, ROUGE-L \cite{lin2004rouge}, and CIDEr \cite{vedantam2015cider}.
The metrics are formulated to compute the word overlapping between predicted responses and ground-truth responses. 

\subsection{Results on DSTC8 Test}
\begin{table*}[h]
\small
\begin{tabular}{llllllllll}
\hline
 \multicolumn{1}{c}{\textbf{Visual}} & \multicolumn{1}{c}{\textbf{Audio}} & \multicolumn{1}{c}{\textbf{BLEU1}} & \multicolumn{1}{c}{\textbf{BLEU2}} & \multicolumn{1}{c}{\textbf{BLEU3}} & \multicolumn{1}{c}{\textbf{BLEU4}} & \multicolumn{1}{c}{\textbf{METEOR}} & \multicolumn{1}{c}{\textbf{ROUGE-L}} & \multicolumn{1}{c}{\textbf{CIDEr}} & \multicolumn{1}{c}{\textbf{Human}} \\
\hline
 ResNext                             & -                                  & 0.724                              & 0.599                              & 0.496                              & 0.414                              & 0.269                               & 0.570                                & 1.101                              & -                                        \\
 I3D(RGB)                            & -                                  & 0.729                              & 0.602                              & \textbf{0.500}                     & 0.417                              & 0.273                               & 0.573                                & 1.108                              & -                                        \\
I3D(Flow)                           & -                                  & 0.724                              & 0.597                              & 0.496                              & 0.413                              & 0.270                               & 0.566                                & 1.110                              & -                                        \\
 -                                   & VGGish                             & 0.730                              & \textbf{0.603}                     & \textbf{0.500}                     & 0.417                              & \textbf{0.274}                      & \textbf{0.576}                       & \textbf{1.113}                     & -                                        \\
ResNext+I3D(RGB)                    & -                                  & 0.695                              & 0.583                              & 0.491                              & 0.416                              & 0.259                               & 0.559                                & 1.087                              & -                                        \\
 ResNext+I3D(Flow)                   & -                                  & 0.696                              & 0.585                              & 0.495                              & \textbf{0.421}                     & 0.261                               & 0.561                                & 1.098                              & 3.609                                    \\
 ResNext                             & VGGish                             & 0.701                              & 0.587                              & 0.494                              & 0.419                              & 0.263                               & 0.564                                & 1.097                              & \textbf{3.612}                           \\
 -                                   & -                                  & \textbf{0.735}                     & \textbf{0.603}                     & 0.497                              & 0.410                              & \textbf{0.274}                      & 0.573                                & 1.108                              & -             \\   
\hline
\end{tabular}
\caption{Result summary on the test dataset in the AVSD benchmark for DSTC8.}
\label{tab:dstc8_result}
\end{table*}
\begin{table*}[h]
\small
\centering
\begin{tabular}{lllllllll}
\hline
\multicolumn{1}{c}{\textbf{Visual}} & \multicolumn{1}{c}{\textbf{Audio}} & \multicolumn{1}{c}{\textbf{BLEU1}} & \multicolumn{1}{c}{\textbf{BLEU2}} & \multicolumn{1}{c}{\textbf{BLEU3}} & \multicolumn{1}{c}{\textbf{BLEU4}} & \multicolumn{1}{c}{\textbf{METEOR}} & \multicolumn{1}{c}{\textbf{ROUGE-L}} & \multicolumn{1}{c}{\textbf{CIDEr}} \\
\hline
 ResNext                             & -                                  & 0.750                              & \textbf{0.619}                     & 0.514                              & 0.427                              & 0.280                               & \textbf{0.580}                       & \textbf{1.189}                     \\
I3D(RGB)                            & -                                  & 0.750                              & 0.617                              & 0.510                              & 0.424                              & 0.282                               & 0.579                                & 1.185                              \\
I3D(Flow)                           & -                                  & 0.750                              & 0.616                              & 0.511                              & 0.427                              & 0.280                               & 0.579                                & 1.188                              \\
 -                                   & VGGish                             & 0.751                              & 0.618                              & 0.511                              & 0.426                              & 0.278                      & \textbf{0.580}                       & 1.186                              \\
 ResNext+I3D(RGB)                    & -                                  & 0.734                              & 0.615                              & 0.517                              & 0.439                              & 0.277                               & 0.574                                & 1.177                              \\
ResNext+I3D(Flow)                   & -                                  & 0.735                              & 0.616                              & \textbf{0.519}                     & \textbf{0.441}                     & 0.277                               & 0.573                                & 1.177                              \\
ResNext                             & VGGish                             & 0.727                              & 0.609                              & 0.515                              & 0.439                              & 0.275                               & 0.574                                & 1.167                              \\
-                                   & -                                  & \textbf{0.752}                     & 0.614                              & 0.507                              & 0.421                              & \textbf{0.283}                      & 0.577                                & 1.185                             \\
\hline
\end{tabular}
\caption{Result summary on the test dataset in AVSD benchmark with test data from the DSTC7.}
\label{tab:dstc7_result}
\end{table*}
\begin{table*}[h]
\small
\centering
\begin{tabular}{llllllll}
\hline
\multicolumn{1}{c}{\textbf{\begin{tabular}[c]{@{}c@{}}Pointer \\ Source Sequence\end{tabular}}} & \multicolumn{1}{c}{\textbf{BLEU1}} & \multicolumn{1}{c}{\textbf{BLEU2}} & \multicolumn{1}{c}{\textbf{BLEU3}} & \multicolumn{1}{c}{\textbf{BLEU4}} & \multicolumn{1}{c}{\textbf{METEOR}} & \multicolumn{1}{c}{\textbf{ROUGE-L}} & \multicolumn{1}{c}{\textbf{CIDEr}} \\
\hline
Summary+Query                                                                                   & \textbf{0.750}                     & \textbf{0.619}                     & \textbf{0.514}                     & \textbf{0.427}                     & \textbf{0.280}                      & \textbf{0.580}                       & \textbf{1.189}                     \\
History+Query                                                                                   & 0.738                              & 0.602                              & 0.494                              & 0.408                              & 0.274                               & 0.568                                & 1.140                              \\
Summary+History+Query                                                                           & 0.739                              & 0.607                              & 0.495                              & 0.407                              & 0.272                               & 0.567                                & 1.142                              \\
Summary                                                                                         & 0.744                              & 0.612                              & 0.505                              & 0.422                              & 0.276                               & 0.576                                & 1.155                              \\
Query                                                                                           & 0.748                              & 0.609                              & 0.499                              & 0.412                              & 0.279                               & 0.573                                & 1.143                              \\
History                                                                                         & 0.738                              & 0.603                              & 0.492                              & 0.401                              & 0.279                               & 0.560                                & 1.061                              \\
None                                                                                            & 0.733                              & 0.597                              & 0.489                              & 0.405                              & 0.269                               & 0.562                                & 1.120     \\                        \hline
\end{tabular}
\caption{Result summary of different model variants of pointer network generator. The results are tested on the test data of AVSD benchmark with test data from the DSTC7.}
\label{tab:ablation_ptr}
\end{table*}
We first report the results on the DSTC8 test dataset.
The results were released by the competition organizer as the ground-truth labels are not publicly accessible.
We submitted different model variants based on the settings of input: (1) text only and (2) text and video. 
In the text-only setting, we remove any visual or audio features and only use text input (including video caption) as input to our model.
In the text-and-video setting, we submitted different versions of our models that either use visual or audio (or both) features combined. 
For visual features, besides ResNext101 as our main visual features, we also utilize the I3D features provided by the organizer. The features are extracted from an I3D \cite{carreira2017quo} model pretrained on the Kinetics dataset. The features have a dimension 2048, the same as ResNext101 features. 

\noindent From Table \ref{tab:dstc8_result}, we noted that the performance among the visual features i.e. ResNext, I3D(RGB), and I3D(Flow), are comparable, especially between I3D(RGB) and I3D(Flow), the differences between objective metrics are minor. 
When only using audio features extracted from VGGish, we note that the performance slightly improves but not significantly as compared to only using visual features.
As compared to the original MTN approach \cite{le-etal-2019-multimodal}, we noted the difference in performance between models that use either visual or audio features is substantially reduced.
We noted similar observations as we compared the difference of performance between models that only use text features and models that use visual features. 
We expect these performance gains come from using pointer generators which can point to tokens in the source sequences i.e. user queries and video summaries. 
Since AVSD is formulated as a generation task with evaluation metrics based on similarity between the generated sentences and the ground truth, we could substantially improve the performance by focusing on the language component of the model. 
We also observed that using a simple ensemble technique could improve the performance, mainly in BLEU-based metrics. 
In this case, the ensemble strategy acts as a regularization factor on the vocabulary distribution of the output, resulting in more semantically correct output sentences.
However, other metrics do not improve or reduce when performing model ensemble.
We obtain the human evaluation scores from the organizers for two of our models. 
Our models achieve human scores of more than 3.6 on a scale of 4 and were ranked top 5 and 6 among all submissions in the AVSD track. 

\subsection{Results on DSTC7 Test}
We also reported the results of the submitted models mentioned above but tested on the test set of DSTC7. 
We note similar observations as ones seen with the test data in DSTC8. 
The overall performance is, however, higher in DSTC7 than DSTC8. 
This reveals that the new test data in DSTC8 is more challenging and the current approach could be further improved.

\subsection{Ablation Analysis}
We evaluated our models with different variants of pointer networks by allowing the models pointing to tokens of different combinations of the text input sequences.
In these experiments, we choose the video-and-test setting and only use visual features extracted from the pretrained ResNext101. 
From Table \ref{tab:ablation_ptr}, we have the following observations.
First, most of the MTN models with our proposed pointer network shows improvement over one that only uses a linear transformation to generate tokens. 
The performance gain is substantial when we allow the models to point to source sequences of video summaries and user queries. 
However, the performance is slightly affected when we use pointer network to point to tokens in dialogue history because user queries and dialogue history typically contain more useful information than dialogue history to generate system responses.
Secondly, when combining different input sequences with multiple pointer networks, the model with the best performance is one that contains pointers to both video summaries and user queries. 
By extending the pointer network to MTN and adopting a dynamic combination of vocabulary distributions among pointers, we can boost the language generation capability of the models and generate better responses.

\subsection{Conclusion}
In this paper, we present our submission AVSD track of the DSTC8 challenge. 
Our submissions achieve competitive performance in both human evaluation and automatic metrics, including BLEU, ROUGE, METEOR, and CIDEr. 
The task is challenging because it involves video information of multiple modalities, including visual and audio information, and it requires strong language modeling capability to generate natural dialogue responses. 
In this work, we focus on the second aspect by adopting pointer networks in generative components. 
Our experiment results show that adopting this technique in video dialogues can improve the quality of the responses.
In the future, we will focus to extend on the first aspect by improving the multimodal reasoning capability between language, visual, and audio features. 

{\small
\bibliographystyle{aaai}
\bibliography{formatting-instructions-latex-2020}
}

\end{document}